\documentclass[a4paper]{article}

\usepackage{INTERSPEECH2018}
\usepackage{listings}
\usepackage{hyperref}
\usepackage{balance}
\usepackage{dblfloatfix}

\title{Conversational Analysis using Utterance-level Attention-based Bidirectional Recurrent Neural Networks}
\name{Chandrakant Bothe, Sven Magg, Cornelius Weber and Stefan Wermter}
\address{
Knowledge Technology, Department of Informatics,
University of Hamburg, \\
Vogt-Koelln-Str. 30, 22527 Hamburg, Germany \\
\url{www.informatik.uni-hamburg.de/WTM/}
\email{\{bothe,magg,weber,wermter\}@informatik.uni-hamburg.de}
}
\begin{document}
\maketitle

\begin{abstract}
Recent approaches for dialogue act recognition have shown that context from preceding utterances is important to classify the subsequent one. 
It was shown that the performance improves rapidly when the context is taken into account. 
We propose an utterance-level attention-based bidirectional recurrent neural network (Utt-Att-BiRNN) model to analyze the importance of preceding utterances to classify the current one. 
In our setup, the BiRNN is given the input set of current and preceding utterances. 
Our model outperforms previous models that use only preceding utterances as context on the used corpus. 
Another contribution of our research is a mechanism to discover the amount of information in each utterance to classify the subsequent one and to show that context-based learning not only improves the performance but also achieves higher confidence in the recognition of dialogue acts. 
We use character- and word-level features to represent the utterances.
The results are presented for character and word feature representations and as an ensemble model of both representations. 
We found that when classifying short utterances, the closest preceding utterances contribute to a higher degree. 

\end{abstract}


\section{Introduction}

Conversational discourse analysis is an important task for natural language understanding and for building a spoken dialogue system. 
A conversation consists of several utterances in a sequence. 
Discourse analysis of the conversation can be conducted by using speech acts where a speech act defines the performative function of an utterance \cite{austin1962things}.
However, speech acts are context-sensitive, where the context provides information for appropriate interpretation of the speech act \cite{sbisa2002speech}. Once the context is taken into account, the question is how many utterances in the context contribute to the current utterance and how do context-utterances affect the interpretation \cite{austin1962things,sbisa2002speech,searle1979,wermter1996learning}?

We attempt to answer these questions in this research. 
We propose an utterance-level attention mechanism using a bidirectional recurrent neural network (Utt-Att-BiRNN) for context-based learning in conversational analysis \cite{GravesBiRnn,SchusterBiRnn,bahdanau2014neural,zhou2016attention}. 
The proposed model is intended to not only model context-based learning but also to analyze the amount of contributing information in the utterances for the dialogue act (DA) recognition task. 
We assess the model performance on the Switchboard Dialogue Act (SwDA) corpus \cite{godfrey1992switchboard}. 

We previously found a significant improvement of using the context-utterances against the simple utterance-level DA classification.
As a result, we now investigate the discourse analysis in a conversation with context-based learning using the Utt-Att-BiRNN model.
We show that context-based learning is important for the conversational analysis improving the performance by 5\% to 8\% accuracy over utterance-level classification.

We also show that the proposed model not only improves the performance but also provides higher confidence in the predicted classes. 
We report the amount of information contributed by the context-utterances.
We experiment with two models: utterance-level model and Utt-Att-BiRNN model.
We discover that there are many instances that were detected wrongly with both models.
These instances are also reported in the results and discussion section along with the samples where the simple utterance-level model fails to predict correctly, as opposed to the Utt-Att-BiRNN model, for the SwDA corpus test set.
With this investigation, we might be able to find ambiguously or wrongly annotated utterances.

\section{Related work}

Previous work in the field of conversational discourse analysis has been attempting to model utterance-level classification of the dialogue acts \cite{stolcke2000dialogue,grau2004dialogue,tavafi2013dialogue,khanpour2016dialogue}. 
However, classifying the DA classes at a single-utterance level might fail when it comes to DA classes where the utterances share similar lexical and syntactic cues (words and phrases) like the \textit{backchannel (b)}, \textit{no-answer (nn)}, \textit{yes-answer (ny)} and \textit{accept/agree (aa)} DA classes. Stolcke et al., 2000 \cite{stolcke2000dialogue} achieve about 71\% of accuracy with hidden Markov models on the SwDA test set.
Many recent works show that context-based learning, which takes the preceding utterances into account, improves the performance of the proposed models to achieve state-of-the-art results \cite{sridhar2008modeling,kalchbrenner2013recurrent,ji2016latent,kumar2017dasl,tzh2017EMNLP,ltAl2017EMNLP,ortega2017neural,Meng2017,lee2016sequential,BOTHE18_525}. 

The context-based learning approach was first proposed to model discourse within a conversation using RNNs. The DA of the current utterance was calculated using the preceding utterances as a context, achieving state-of-the-art results of about 74\% of accuracy on SwDA \cite{kalchbrenner2013recurrent,ortega2017neural}. Kalchbrenner and Blunsom, 2013 \cite{kalchbrenner2013recurrent} represent the utterance as a compressed vector of word embeddings using convolutional neural networks (CNN) and use these utterance representations to model discourse within a conversation using RNNs. 
Lee and Dernoncourt, 2016 \cite{lee2016sequential} also use recent techniques such as RNNs and CNNs with word-level feature embeddings and achieve about 73\% of accuracy. Ortega and Vu, 2017 \cite{ortega2017neural} also use CNNs and RNNs and achieve about 74\% of accuracy.

In another line of research, the context-based learning approach processes the whole set of utterances in a conversation, where the model can see past and future utterances to calculate the DA of the current utterance \cite{ji2016latent,kumar2017dasl}. 
Ji et al. 2016 \cite{ji2016latent} use discourse annotation for the word-level language modelling on the SwDA corpus and achieve about 77\% of accuracy but also highlight a limitation that this approach is not scalable to large data.
On the other hand, this work suggests that a domain-independent language model which is trained on big data might be a solution.
In some approaches, a hierarchical convolutional and recurrent neural encoder model are used to learn utterance representations by processing a whole conversation \cite{kumar2017dasl,ltAl2017EMNLP}.
The utterance representations are further used to classify DA classes using the conditional random field (CRF) as a linear classifier.
However, these models might fail in a dialogue system where one can perceive the past utterances, but cannot see future ones. 

In a dialogue system for example in human-machine interaction, one can only perceive the preceding utterance as a context but does not know the upcoming utterances.
The DA corpus is also annotated by looking at the preceding utterances.
Therefore, we use a context-based learning approach where only preceding utterances are considered and regard the 73.9\% accuracy \cite{kalchbrenner2013recurrent,ortega2017neural} on the SwDA corpus as a current state-of-the-art result for this particular task.

\section{Experimental setup}

\subsection{Dataset}

Discourse analysis is a very important task in the field of natural language processing and hence there are many dialogue act corpora available \cite{serban2015survey}. We use the Switchboard Dialogue Act (SwDA\footnote{\url{https://github.com/cgpotts/swda}}) corpus which is annotated with the Dialogue Act Markup in Several Layers (DAMSL) tag set \cite{godfrey1992switchboard,shribergSwitch}. SwDA is annotated with 42 DA classes. 
The corpus consists of 1,115 conversations (196,258 utterances) in the training and 19 conversations (4,186 utterances) in the test set \cite{stolcke2000dialogue,kalchbrenner2013recurrent}.

\subsection{Utterance representations}
\label{sec:utt_rep}

We represent each utterance with two different speech-language features: characters and words. 

\textbf{Character representations:} The character-level utterance is encoded with a pre-trained character-level language model (LM\footnote{\url{https://github.com/openai/generating-reviews-discovering-sentiment}}) \cite{Radford2017LearningSentiment}. 
This model consists of a single multiplicative long-short-term memory (mLSTM) network \cite{krause2016multiplicative} layer with 4,096 hidden units. 
The mLSTM is composed of an LSTM and a multiplicative RNN and considers each possible input in a recurrent transition function.
It is trained as a character language model on $\sim$80 million Amazon product reviews \cite{Radford2017LearningSentiment}.
We sequentially input the characters of an utterance to the mLSTM and get the hidden vector obtained after the last character and also average the states over all characters in the utterance.
We use the average feature vector representations for each utterance in the experiments as it was shown that the average vector over all characters in the utterance works better for dialogue act recognition \cite{BOTHE18_525} and for emotion detection \cite{labothe2017WASSA2017}. 

\textbf{Word representations:} Word-level features are important for analyzing the short sentences in a conversation. We use the word-embeddings distributed as part of ConceptNet 5.5\footnote{\url{https://github.com/commonsense/conceptnet-numberbatch}} as it is designed to represent the general knowledge involved
in understanding language and allows the application to better understand
the meanings behind the words people use \cite{speer2017conceptnet}. It has a knowledge graph that connects words and phrases of natural language with labelled edges. 
The embedding dimension is 300 and averaged over all tokens in the utterance. These embeddings provide the out-of-vocabulary instance rate close to 10 percent and mostly for infrequent words.

\subsection{Utterance-level attention-based BiRNN}

First, we present our baseline model as shown in Figure \ref{fig:proposed_model}(a), it is a simple utterance-level classifier which classifies the utterances with their respective labels (dialogue acts) using a simple feed-forward neural network with backpropagation.  
The Utt-Att-BiRNN model is shown in Figure \ref{fig:proposed_model}(b), for which the main components are the bidirectional recurrent neural network (\textit{BiRNN}) and \textit{Attention} mechanism.

\subsubsection{Bidirectional recurrent neural network} 

A BiRNN is an extended form of an unidirectional RNN \cite{elman1990finding}, introducing one extra hidden layer \cite{GravesBiRnn,SchusterBiRnn}. The hidden to hidden layer connections flow into the opposite temporal direction. 
The model provides forward and backward states with corresponding directions of the hidden layers, as shown in Figure \ref{fig:proposed_model}(b), and the final result is calculated as follows:
\begin{equation}
h_{t}^{f}  = f\left(W_{h}^{f}h_{t-1}^{f}+W_{u}^{f}u_{t} + b_{h}^{f}\right)
\end{equation} 
\begin{equation}
h_{t}^{b}  = f\left(W_{h}^{b}h_{t-1}^{b}+W_{u}^{b}u_{t} + b_{h}^{b}\right)
\end{equation} 
\begin{equation}
P(y_{t} | \{u_{t}, u_{t-1},...u_{t-n}\}) = g\left(W_{y}^{f}h_{t}^{f} + W_{y}^{b}h_{t}^{b} + b_{y}\right)
\end{equation} 
where $n$ is the number of utterances in the context for time instance $t$.
$W$ and $h$ are the corresponding weight matrices and hidden vectors, where the superscripts $f$ and $b$ represent forward and backward hidden layer directions respectively. 
In our scenario, we want the model to learn the context, thus the input consists of the current utterance and the preceding context. 
If we use a unidirectional RNN model, there might be a chance that the model becomes more attentive to the current utterance only, as sequential information is compressed to the final state.
The bidirectional RNN model, on the other hand, exploits the information in all given input utterances by looking back and forth through them. 
Therefore, our goal is to treat all utterances equally and learn how much each contribute to the final result.

\subsubsection{Attention mechanism}
\label{sec:utt_att}

\begin{figure}[t]
  \centering
  \includegraphics[width=\linewidth]{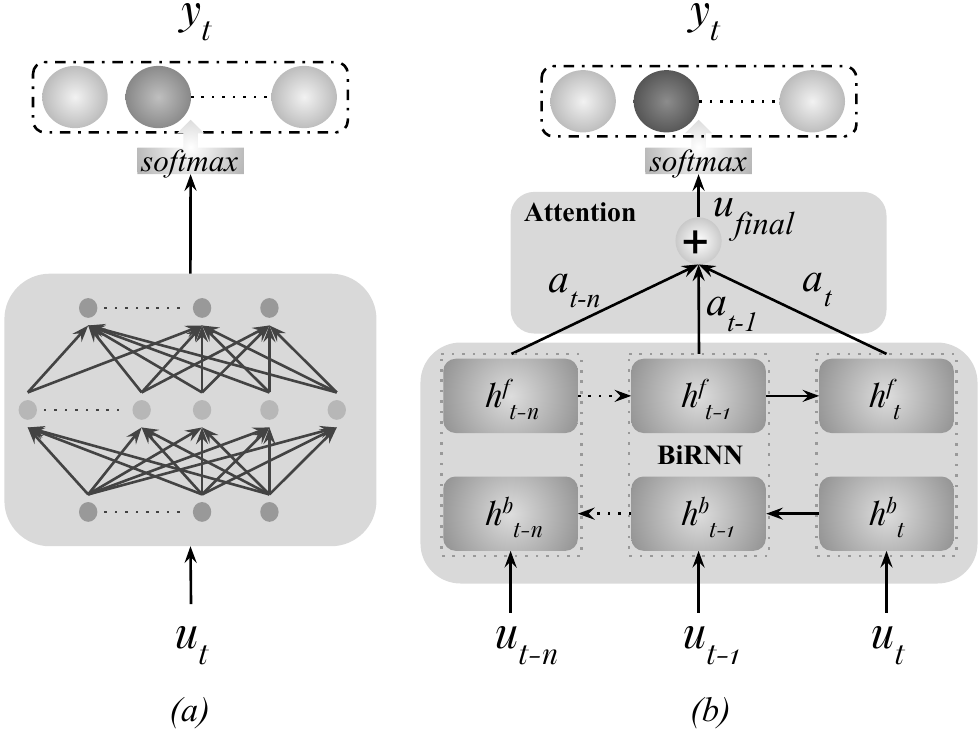}
  \caption{(a) Our baseline model, (b) Utt-Att-BiRNN model.}
  \label{fig:proposed_model}
\end{figure}

Attention mechanism is loosely based on visual attention found in humans, and broadly used in image recognition and tracking \cite{NIPS2010_4089,denil2012learning}. 
But recently, attention mechanism with RNNs are being used for several natural language processing tasks, such as machine translation and comprehension, speech recognition \cite{bahdanau2014neural,vinyals2015grammar,chorowski2015attention}. 
We propose the attention mechanism to compute the contribution weights of the utterances for predicting the corresponding class. 
Given the number ($n$) of preceding utterances in an input sequence $u =  \{u_{t}, u_{t-1},...u_{t-n}\}$, the BiRNN provides the respective hidden vectors $h =  \{h_{t}, h_{t-1},...h_{t-n}\}$. 
The attention layer computes the weights $a = \{a_{t}, a_{t-1},...a_{t-n}\}$ as the contribution for every corresponding input utterance in $u$ using the respective hidden representations $h$, as depicted in Figure \ref{fig:proposed_model}(b). 
Hence, the final utterance representation $u_{final}$ of the utterance sequence in $u$ is formed by a weighted sum of $h$ and $a$:
\begin{equation}
m  = tanh\left(W_{h}h\right)
\end{equation} 
\begin{equation}
a  = softmax\left(W_{m}^{T}m\right)
\end{equation} 
\begin{equation}
u_{final} = tanh\left(ha^{T}\right)
\end{equation} 
where $W$ is a trained parameter while $W^{T}$ being its transpose. We use the $softmax$ function to compute the weights which provides $\sum_{n=0}^{n} a_{t-n} = 1$. It is important for the utterance-level attention mechanism that we normalize $a$ to interpret the amount of contribution for each utterance in $u$. 

\begin{table}[b]
  \caption{Accuracies (in \%) on the SwDA test set, baseline with no context (NC) and Utt-Att-BiRNN model with context (WC)}
  \label{tab:accOnSwDA}
  \centering
  \begin{tabular}{ lll }
    \toprule
    \textbf{Models}       & \textbf{NC}  & \textbf{WC}  \\
\midrule
    \textbf{Prior related work}    &    &  \\
    Most common class baseline        & $31.50$~~~    &  \\
    Stolcke et al., 2000 \cite{stolcke2000dialogue} & $71.00$~~~ &    \\
    Kalchbrenner and Blunsom, 2013 \cite{kalchbrenner2013recurrent} & & $73.90$~~~ \\
    Lee and Dernoncourt, 2016 \cite{lee2016sequential} & & $73.10$~~~ \\
    Ortega and Vu, 2017 \cite{ortega2017neural} & & $73.80$~~~ \\
    \textbf{Our work}    &    &  \\
    Character LM rep. &   $71.84$~~~  & $76.47$~~~ \\
    Word-embeddings mean rep.  & $71.73$~~~ & $75.43$~~~ \\
    Concatenated rep.  & $70.83$~~~  &  $76.15$~~~  \\
    Average char-word-level predictions  & $71.85$~~~ & $76.84$~~~ \\
 
    Average char-word-level \&        &    &    \\
    concatenated rep. predictions & $71.97$~~~ & \textbf{$77.42$~~~} \\
    \bottomrule
  \end{tabular}
\end{table}

\subsubsection{Training the model} 

In the baseline model and the Utt-Att-BiRNN model settings, we use a $softmax$ function to predict a discrete set of classes $y$ on top of the learned $u_{final}$ representations. 
We use a set of 5 utterances in $u$, with the current utterance and 4 utterances in the context. 
A similar study performed in \cite{BOTHE18_525} shows the effect of the number of utterances in the context. It was shown that three utterances provide sufficient context, however, we use four context-utterances to provide a large enough window for bidirectional exploration by the RNN, hence $n=4$.

In all learning cases, we minimize the categorical cross-entropy as we have multiple classes in the DA recognition task. 
For the baseline model, we use 2 hidden layers with 300 and 100 hidden units respectively. For the proposed model, we use 64 hidden units with the dropout regularizer \cite{hinton2012improving} in the BiRNN hidden layer. As a result, we get 128 hidden units as a concatenation of the $h_{t}^{f}$ and $h_{t}^{b}$ hidden units. 
These are the only parameters determined empirically for the classification tasks but all other parameters were learned during training.

The Adam optimizer \cite{kingma2014adam} was used with an initial learning rate 1\textit{e}-4, which decays during training.
Early stopping was used to avoid over-fitting of the network, and 15\% of training samples were used for validation. 
We wait for at least 5 iterations over which the accuracy on the validation set does not improve.
Typically, both models, baseline and Utt-Att-BiRNN, took about 20 to 30 interations.

\section{Results and discussion}

The baseline and Utt-Att-BiRNN models are trained and tested using both the utterance representations explained in Section \ref{sec:utt_rep}. 
We report the accuracies on a test set of the SwDA corpus in Table \ref{tab:accOnSwDA}. 
Character LM and word-embeddings mean utterance representations perform quite well for this task. 
Surprisingly, the word-embeddings mean representations of the utterances used from the ConceptNet seem to show good results given the fact of the low dimensionality of the embeddings (300) compared to character LM (4096) size.

\begin{table}[b]
\centering
\caption{The test samples from the SwDA corpus where both classifiers, simple utterance-level and Utt-Att-BiRNN, failed to correctly predict classes (the majority classes, Statement-non-opinion (\textit{sd}) and Statement-opinion (\textit{sv}), are reported here). Where Num is a number of samples, GT stands for ground truth, and pct. for percentage.}
\label{tab:faliurepredictions}
\begin{tabular}{llllll}
\toprule
\textbf{GT} & \textbf{NC}  &  \textbf{WC} & \textbf{Num} & \textbf{pct.} & \textbf{Example of utts} \\
\midrule
\textit{sv}   &  \textit{sd}   &  \textit{sd}   & 198   &  4.73  & Uh, the problem is here \\
     &       &       &       &        & But they don't have \\
\vspace{0.3cm}
     &       &       &       &        & We're hearing the same \\
\textit{sd}   &  \textit{sv}   &  \textit{sv}   & 51    &  1.22  & They're certainly legal, \\
     &       &       &       &        & Real long legs, \\
     &       &       &       &        & And time consuming, \\

\bottomrule
\end{tabular}
\end{table}

We also experiment with a combined model of these representations in two ways: first by concatenating both representations and use them as an input, and second by averaging the output predictions from both models. 
Averaging the predictions has shown the best results, and we even found that the average of prediction of models trained with character LM, word-embeddings mean, and concatenated representations give the best of the performance.  
We can see that context-based learning shows a performance improvement of about 5\% on this discourse analysis task.

\begin{figure*}[t]
  \centering
  \includegraphics[width=\linewidth]{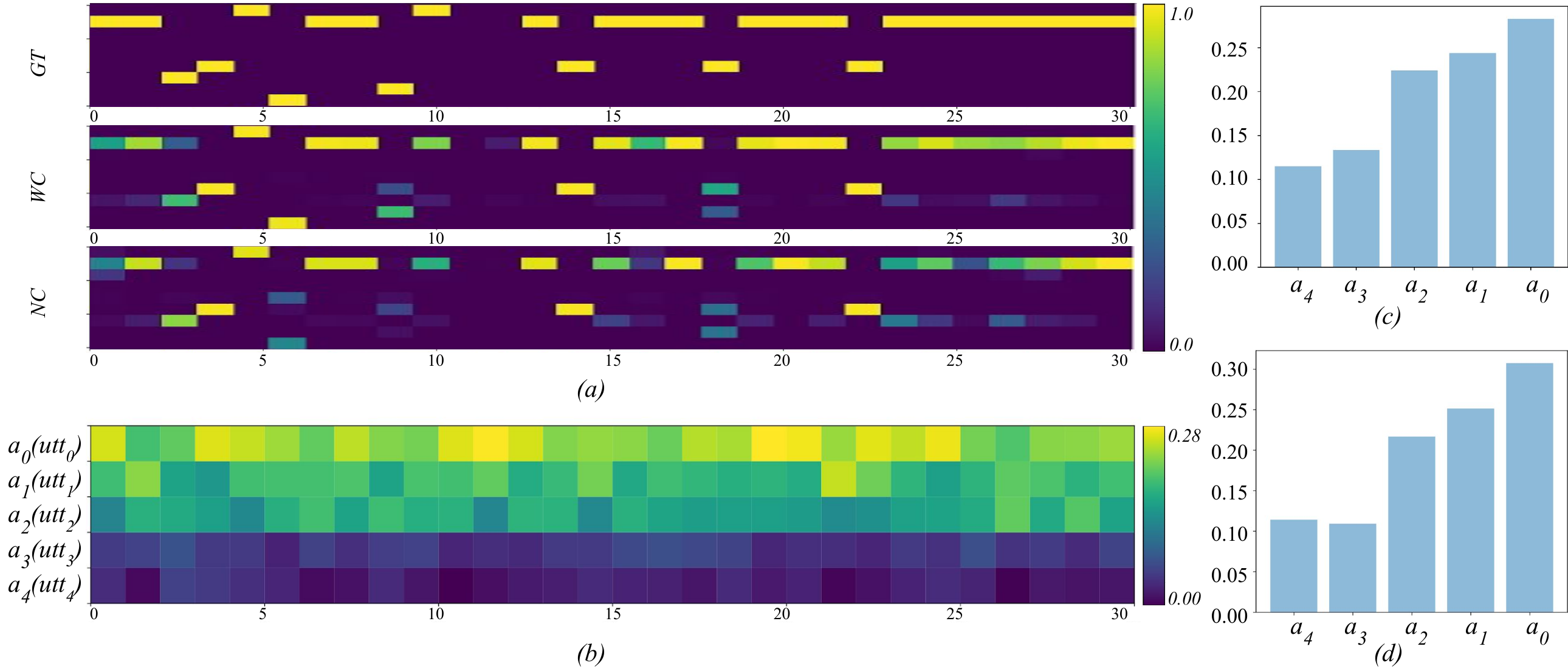}
  \caption{Effectiveness of the context. 
  (a) Prediction confidence for a batch of 30 sets of utterances: the first row is the ground truth (GT), the second row the predictions with context (WC), and the third row the predictions with no context (NC). 
  We show only 8 of the 42 classes for simplicity on the y-axis and the set of utterances on the x-axis. 
  (b) The contribution of utterances $utt_{0}, utt_{1},...utt_{4}$ as the attention weights $a_{0}, a_{1},...a_{4}$. 
  (c) The average weight of utterances and 
  (d) in addition averaged over 10 runs to show robustness.} 
  \label{fig:completeFigure}
\end{figure*}

We examined the SwDA corpus test set and found that there are many instances that were predicted wrongly with both models. 
The dominant DA classes in the SwDA corpus are Statement-non-opinion (\textit{sd}) and Statement-opinion (\textit{sv}).
Table \ref{tab:faliurepredictions} shows the number of samples (\textit{Num}) and the percentage (\textit{pct.}) out of 4,186 utterances. 
The examples of utterances show that it might be difficult for humans to identify the correct DA class. It shows the ambiguity in the two DA classes \textit{sd} and \textit{sv}, which accounts about 6\% of accuracy reduction for both of the models only with these two classes.
We also show the effectiveness of the pragmatic model which predicts the correct class when the context is important, see Table \ref{tab:correctpredictions}.
For example, if the utterances like "Yes", "Yeah" etc. are followed by Yes-No Question (\textit{qy}), the probability that the second utterance belongs to Yes-Answer (\textit{ny}) is higher than being in Backchannel (\textit{b}) or Abandoned (\textit{\%}). 
Similar utterances to the \textit{ny} class are used in the Agree/Accept (\textit{aa}) class, but they are usually followed by \textit{sv}, \textit{sd}, \textit{b}, or some other classes. 
In total, we found 330 samples which constitute around 7.88\% of the samples that were correctly recognized by the Utt-Att-BiRNN model but not by the utterance-level model.

\begin{table}[b]
\centering
\caption{The test samples from the SwDA corpus where the Utt-Att-BiRNN model correctly predict as opposed to the simple utterance-level classifier.}
\label{tab:correctpredictions}
\begin{tabular}{lllll}
\toprule
\textbf{GT} & \textbf{NC} & \textbf{WC} & \textbf{Num} & \textbf{pct.} \\
\midrule
\textit{ny}  &  \textit{b}    &  \textit{ny}   & 33    &  0.79  \\
\textit{aa}  &  \textit{b}    &  \textit{aa}   & 29    &  0.69  \\
\textit{aa}  &  \textit{sd}   &  \textit{aa}   & 12    &  0.28  \\
\textit{b}   &  \textit{aa}   &  \textit{b}    & 23    &  0.55  \\
\textit{b}   &  \textit{\%}   &  \textit{b}    & 16    &  0.38  \\
\bottomrule
\end{tabular}
\end{table}

However, we also found that the prediction confidence of the Utt-Att-BiRNN model is higher than the utterance-level classifier. 
Figure \ref{fig:completeFigure}(a) shows three rows for 30 batches of utterance sets in the DA recognition task: first ground truth, second the predictions of the Utt-Att-BiRNN model, and third the predictions of the utterance-level classifier.
The predictions of the Utt-Att-BiRNN model show higher confidence when compared to the predictions of the utterance-level model.

With the help of Utt-Att-BiRNN model we also computed the amount of contribution of the context utterances.
As discussed in Section \ref{sec:utt_att}, the attention weights ($a_{t}, a_{t-1},...a_{t-n}$) can be interpreted as the contribution of the utterances, as the $u_{final}$ of the utterance sequence in $u$ is formed by a weighted sum of $h$ and $a$.
Figure \ref{fig:completeFigure}(b) shows the attention weights ($a_{0}, a_{1},...a_{4}$) that represent the contribution of the corresponding utterances ($utt_{0}, utt_{1},...utt_{4}$).
It is clear that the current utterance $utt_{0}$ contributes more than others, however, the closest preceding utterances seem to contribute substantially.
In Figure \ref{fig:completeFigure}(c) and \ref{fig:completeFigure}(d), we can see the average of the weights for the corresponding utterances.

\section{Conclusions and future research}

In this article, we have presented the Utt-Att-BiRNN model for conversational analysis.
We demonstrated that our model allows not only to model context-based pragmatic learning but also to compute the amount of information used from the context. 
Our model achieves a state-of-the-art result on the SwDA corpus of about 77\% of accuracy, using only preceding utterances in the context.
We showed that our model correctly predicted a significant number of the instances on a DA recognition task. 
We also show that the context-based learning approach shows higher confidence on the classification task compared to simple utterance-level classification.
We have investigated different aspects of the conversational analysis and tested on an important task: dialogue act recognition.

In this research, we only analyzed the utterance representations based on transcripts. 
However, we plan to use audio features in addition which could provide better representations.
Furthermore, it would also help to analyze and mitigate the influence of transcription errors.
We investigated the DA annotations by reviewing the predictions of different models, but we could extend it to find out a reliable metric to assess the model performance.
\balance

\section{Acknowledgements}

This project has received funding from the European Union's Horizon 2020 framework programme for research and innovation under the Marie Sklodowska-Curie Grant Agreement No. 642667 (SECURE).

\bibliographystyle{IEEEtran}
\bibliography{mybib}

\end{document}